\documentclass[10pt, a4paper]{article}
\usepackage{lrec2022} %
\usepackage{multibib}
\newcites{languageresource}{Language Resources}
\usepackage{graphicx}
\usepackage{tabularx}
\usepackage{soul}
\usepackage{titlesec}
\titleformat{\section}{\normalfont\large\bfseries\center}{\thesection.}{1em}{}
\titleformat{\subsection}{\normalfont\SmallTitleFont\bfseries\raggedright}{\thesubsection.}{1em}{}
\titleformat{\subsubsection}{\normalfont\normalsize\bfseries\raggedright}{\thesubsubsection.}{1em}{}
\renewcommand\thesection{\arabic{section}}
\renewcommand\thesubsection{\thesection.\arabic{subsection}}
\renewcommand\thesubsubsection{\thesubsection.\arabic{subsubsection}}

\usepackage{epstopdf}
\usepackage[utf8]{inputenc}

\usepackage{hyperref}
\usepackage{xstring}

\usepackage{color}

\usepackage{booktabs}
\usepackage{enumitem}

\usepackage{xstring}

\title{Please, Don't Forget the Difference and the Confidence Interval when Seeking for the State-of-the-Art Status}

\name{Yves Bestgen} 

\address{Laboratoire d'analyse statistique des textes\\
  Universit\'e catholique de Louvain \\
  Place Cardinal Mercier, 10 1348 Louvain-la-Neuve, Belgium \\
         yves.bestgen@uclouvain.be\\}

\abstract{
This paper argues for the widest possible use of bootstrap confidence intervals for comparing NLP system performances instead of the state-of-the-art status (SOTA) and statistical significance testing. Their main benefits are to draw attention to the difference in performance between two systems and to help assessing the degree of superiority of one system over another. Two cases studies, one comparing several systems and the other based on a K-fold cross-validation procedure, illustrate these benefits. A python module for obtaining these confidence intervals as well as a second function implementing the Fisher-Pitman test for paired samples are freely available on PyPi.
 \\ \newline  }

\begin{document}

\maketitleabstract

\section{Introduction}

With the explosion of NLP studies, more and more systems are developed and evaluated using more and more shared tasks and freely available datasets. Comparing systems has therefore become an essential question \cite{Berg-Kirkpatrick2012,Dodge2019}. Very often, it boils down to ranking systems by performance like in leaderboards or determining whether the new system scores better than the previous better one, the widely discussed "glorification of benchmarks" and state-of-the-art status (SOTA)  \cite{dehghani2021benchmark}. The major advantage of this ranking criterion is that it only requires the performance level of each system.

Using simple ranking has been criticized because the superiority of one system over another may result from chance alone rather than from any real difference \cite{Koehn2004}. The answer is obviously statistical significance testing, whose function is to reject the null hypothesis according to which there is no difference in the population to which the evaluation set belong. Unlike SOTA, it requires access to the predictions of the models to be compared. Yet, it has been widely emphasized, but more often in other fields than NLP \cite{Cohen1994,Ioannidis2019,Kim2019,Rao2016,Schmidt1997}, that null-hypothesis significance testing presents many problems. When comparing NLP systems, one of the most important issues is that a statistical test leads to dichotomizing the results of a study in significant vs. not significant, while statistically significant does not mean important. For most statistical tests, any difference, however small it may be and therefore as unimportant as it is, can be declared significant with enough data. This is for example the case when a chi-square test is (wrongly) used to compare the frequency of words or lexical bundles in large corpora: the smallest difference in frequency gives rise to a statistically highly significant result \cite{Cohen1994,Bestgen2014InadequacyOT,doi:10.1080/09296174.2019.1566975}.

For these reasons, statisticians and researchers in many disciplines have strongly recommended reporting an effect size measure to help evaluating the importance of the result \cite{Claridge-Chang2016,Cohen1994,Masson2003,Saville1990,Wasserstein2019}. When comparing two systems, the most obvious measure of importance is the true difference in performance between them, i.e. the difference in performance that would be observed if the systems were compared on all data they could process. Unfortunately, but obviously, this value is unknown. The observed difference is the best estimate available, but it can be more or less far from the true difference. This is where the confidence interval (CI) comes into play. 

\begin{figure*}
\centering
\includegraphics[width=.89\linewidth]{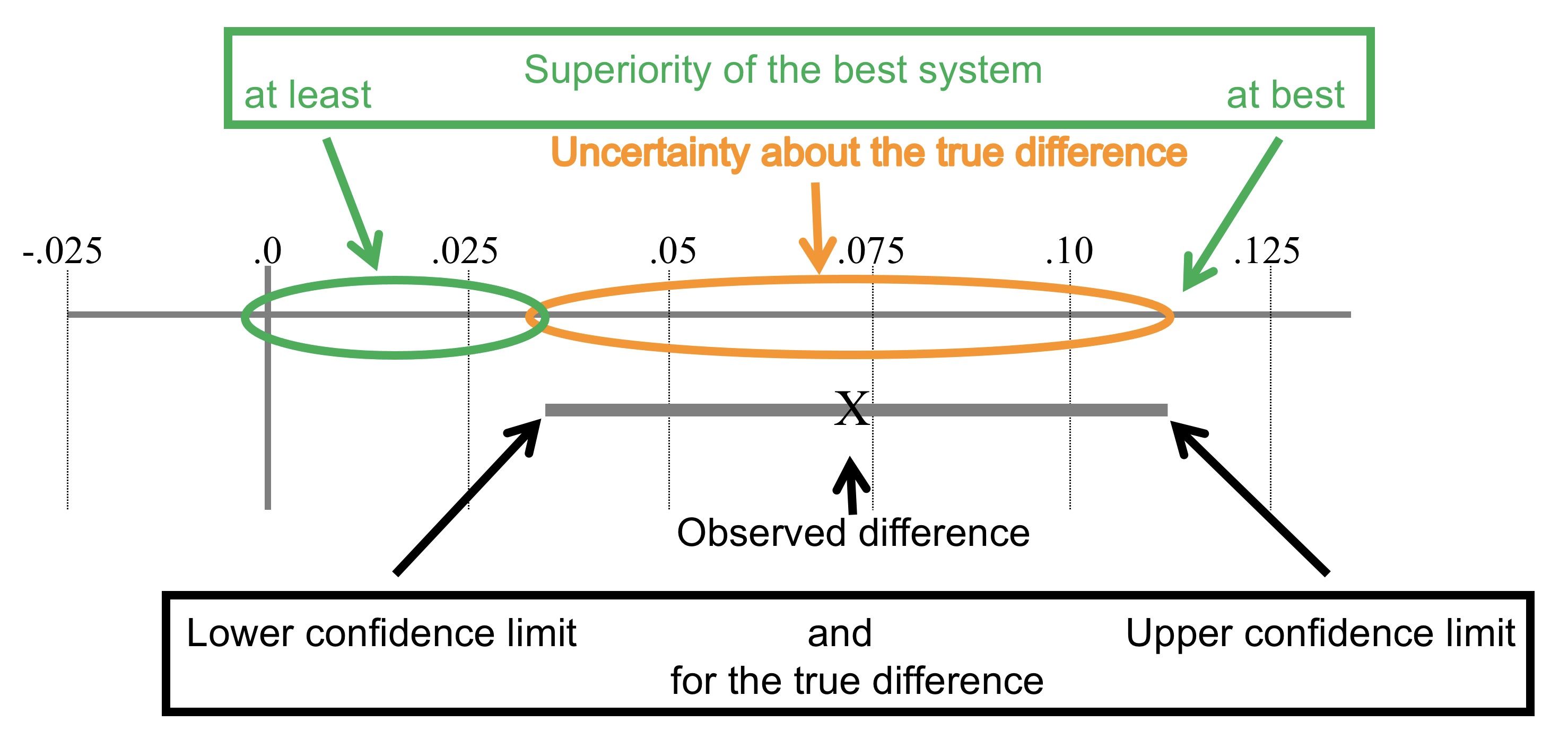}
\caption{Confidence interval for a difference in accuracy between two systems}
\end{figure*}

Conceptually close to the hypothesis test\footnote{A CI at 95\% provides a test of significance at the .05 threshold since the true difference can be declared different from 0 if the CI does not include this value. However, it leads to dichotomizing the results as significant or not, the opposite of what a CI is done for \cite{Wasserstein2019}.}, it is, on the basis of the available data, the best guess, at some level of confidence (e.g. 95\%), of an interval that contains the true difference between the two compared systems  \cite{Ismay2019}. More formally, the interpretation of a CI at 95\% is as follows: if a large number of samples are extracted from the population in question and a CI is calculated for each of these samples, 95\% of them should include the population parameter \cite{Dix2020}.

Figure 1 shows a graphical representation of a confidence interval. The length of the interval informs about the uncertainty associated with the estimate of the true difference and the size of the gap between the confidence limits and 0 helps to assess the degree of superiority of one system over another. This paper argues for the widest possible use of CIs in NLP system comparison.

Calculating CIs for comparing systems is not new in NLP. For example, \newcite{Koehn2004}, \newcite{Zhang2004} and \newcite{Graham2014} proposed to use bootstrap-style CIs to compare machine translation systems while \newcite{Bisani2004} (see also \newcite{Liu2019}) made the same proposal to compare the word error rates of automatic speech recognition systems. However, these studies have emphasized the use of CIs to determine whether a difference is statistically significant and not to deepen the interpretation of the differences in performance between systems. \newcite{Soboroff2014} recommended the use of CIs in IR and compared several computation techniques to obtain them. He focused on finding confidence limits for the performance of a system but suggested that they could be useful for comparing several systems. More recently, \newcite{Lucic2018} used bootstrap CIs to get a better view on a system performance than that provided by its best result.
 
To my knowledge, the closest study is that of \newcite{Berrar2013} who compared the advantages and disadvantages of using significance tests and CIs for the comparison of classifiers. If the arguments presented here are very close to theirs, the proposed approach is much more general by making use of bootstrap CIs.

\section{Using Bootstrap CI for Comparing Systems}

Since many different performance measures are used in NLP and since the statistical properties of these measures are frequently poorly known, \newcite{Berg-Kirkpatrick2012} and \newcite{Koehn2004} have recommended to use a resampling approach such as paired bootstrap because it can be applied to any performance metric. Bootstrap CI calculation works as follow \cite{Diciccio1996}. 

Given a evaluation set of $N$ instances for which the predictions of the two systems have been obtained, by means of some cross-validation procedure, 

\newlist{inparaenum}{enumerate}{2}%
\setlist[inparaenum]{nosep}%
\setlist[inparaenum,1]{label=\arabic*.}%
\setlist[inparaenum,2]{label=\arabic{inparaenumi}\emph{\alph*})}%

\begin{figure*}
\centering
\includegraphics[width=.99\linewidth]{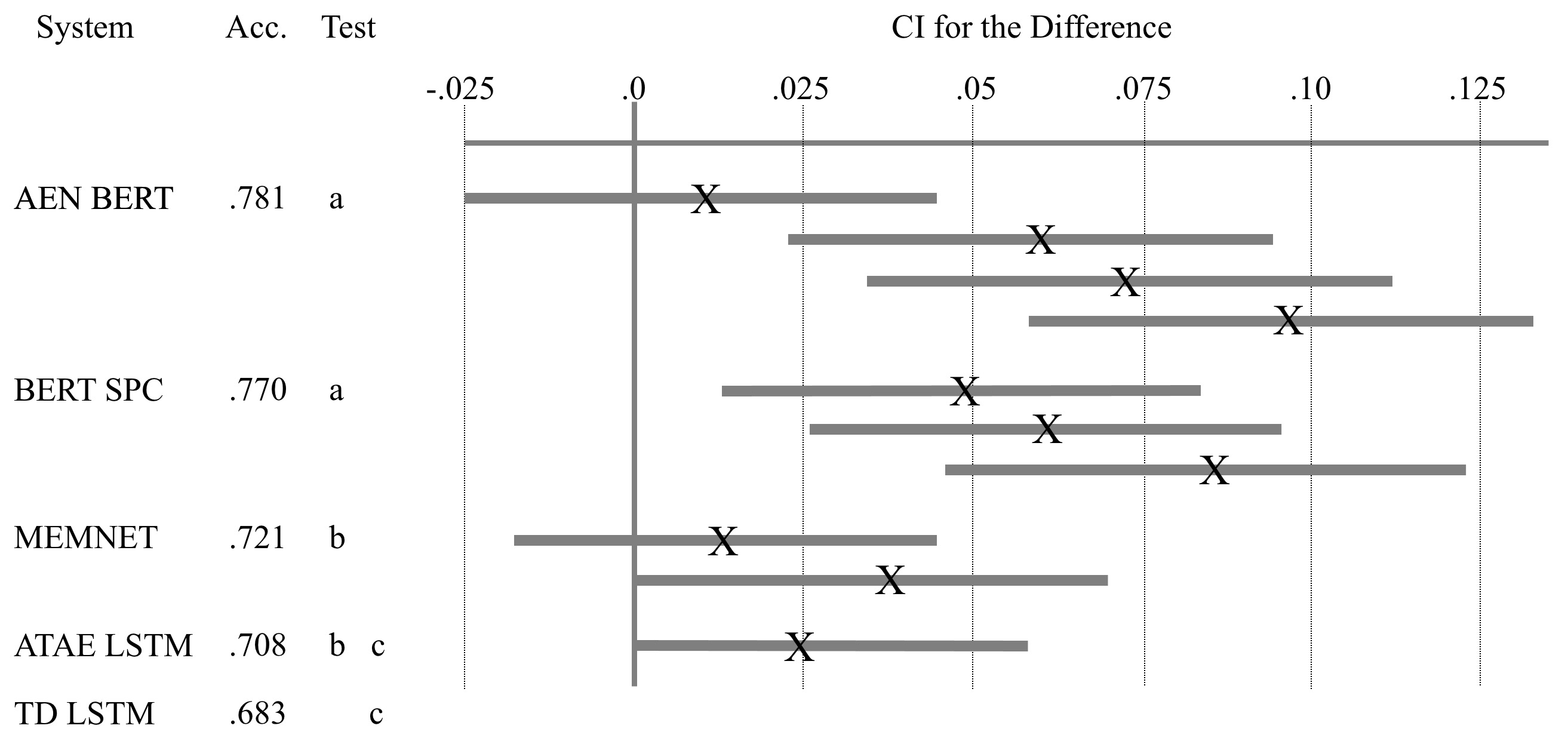}
\caption{Results of the different ways of comparing systems for ABSA. The systems are ordered vertically according to their accuracy, which is given to the right of the name. The letters (a, b and c) indicate the conclusions of the resampling test ($\alpha = .05$), the performances of two systems having the same letter are not significantly different. The CIs at 95\% are given for the difference between each system and those ranked lower, presented in rank order. Thus, the upper CI is for the difference between AEN~BERT and BERT~SPC and the next one is for for the difference between AEN~BERT and MEMNET. Only the first CI of a system is aligned with its name. The \textit{X}s indicate the observed differences.}
\end{figure*}

\begin{inparaenum}
  \item Compute the observed difference in performance of the two systems on the whole evaluation set, the best available estimate of the true difference.
  \item Repeat the following three steps a large number of times\footnote{For a 95\% CI, 1000 to 5000 bootstrap samples are considered as enough \cite{Hesterberg2003}.} to get the bootstrap distribution of the difference.
  \begin{inparaenum}
    \item Randomly sample with replacement $N$ instances from the evaluation set to get a bootstrap sample.
    \item Calculate the performances of each system on this bootstrap sample.
    \item Calculate the difference between these two performances.
  \end{inparaenum}
  \item Use the bootstrap distribution to obtain the CI for the difference.
\end{inparaenum}
Different ways of calculating this CI have been proposed such as the percentile method and the bias corrected method with acceleration (BCa), which is considered to be quite accurate, even for skewed distribution, unless the sample size of the evaluation set is very small \cite{Hayden2019,Hesterberg2003}.

It is noteworthy that, like a significance test, CIs are affected by the sample size of the evaluation set. \textit{Everything else being equal}, larger sample sizes produce narrower CIs. However, the change in interpretation is very different compared to that of a significance test. A very large sample will produce a very small p-value and the researcher can claim to be very confident about the existence of a real difference \textit{whatever its size}. The larger the sample, the smaller the CI and therefore the more confident the researcher can be about the true difference estimate, but this estimate will be identical. If the observed difference is unimportant, it will remain unimportant \cite{Kim2019}.

A python module for obtaining these CIs is freely available on PyPi and can be installed with \textit{pip install BootCI} \cite{BootCI}. Details about to use it is available at (\url{https://github.com/ybestgen/BootCIExpli/}. It is based on the \textit{bootstrap\_interval} module by Alexander Neshitov  (\url{https://pypi.org/project/bootstrap-interval/}, MIT Licence,  \cite{Nesh}) which implements the bootstrap CI for one sample described in \newcite{Diciccio1996}. It was adapted to the comparison of paired samples. The performance measure can be any measure provided by a standard python library or defined by the user. To facilitate the comparison of the CI approach with the significance test one, a second function implementing the Fisher-Pitman test for paired samples \cite{Berry2002,BESCHI17,Neuhauser2004} is also provided in the \textit{BootCI} module. As the bootstrap CI, this test is based on a resampling procedure.
 
\section{Two Case Studies}
The rest of this paper presents two case studies in the field of emotion mining which illustrate the use of CIs for comparing NLP systems. Explanations, code and data for reproducing these cases studies are freely available. They can be downloaded from \url{https://github.com/ybestgen/BootCIRealData}. 

The reported CIs were obtained with the BCa technique, as recommended in the previous section, using 10,000 samples. The same number of samples were used for the Fisher-Pitman tests. In all the analyses, $\alpha$, the probability of wrongly rejecting the null hypothesis, was set to the usual .05 and the confidence to $(1 - \alpha)$. As CIs are interpreted as providing uncertainty bounds on an estimate and not as significance tests, there is no reason to use a Bonferroni-type procedure whose function is to control the overall error rate in multiple comparisons \cite{Saville1990,Soboroff2014}. To allow a fair comparison, a Bonferroni procedure was also not applied to the Fisher-Pitman tests.

\subsection{Study 1:  Song et al. (2019) on ABSA}
The first case study is based on the Laptop review dataset of the SemEval 2014 Task 4 on aspect level sentiment classification (ABSA) \cite{Pontiki2014,Song2019}. This task aims at identifying the sentiment expressed towards an aspect in a sentence, using three sentiment polarities: positive, neutral and negative. The metrics to evaluate the performance was Accuracy (but some researchers reported also Macro-F1). I obtained the predictions on the challenge evaluation set by five systems using the open source ABSA-PyTorch implementation\footnote{ABSA-PyTorch implements more systems than the five included here, but, as this software was designed to compare their model sizes, the parameters used by the different procedures are not optimized and the results I have obtained are sometimes significantly lower than those published in the original studies. This does not pose a problem for the case study, but it seemed preferable to compare only systems whose performances are relatively close to the published values.} provided by \newcite{Song2019} and available at \url{https://github.com/songyouwei/ABSA-PyTorch} \cite{ABSA}: AEN~BERT and BERT~SPC of \newcite{Song2019}, MemNet \cite{tang-etal-2016-aspect}, ATAE~LSTM \cite{wang-etal-2016-attention} and TD~LSTM \cite{tang-etal-2016-effective}.
 
Figure 2 shows the five systems, their accuracy on the evaluation set, the conclusions of the Fisher-Pitman test and the CIs for the differences between the systems.
If the leaderboard approach is used, it is enough to look at the order of the systems; no statistical analysis is necessary. The significance tests indicate that AEN~BERT and BERT~SPC outperform the other three systems and that MEMNET outperforms TD~LSTM. 

The analysis of the CIs leads to more nuanced and richer conclusions. As an example, the difference observed, the best estimate of the true difference, between BERT~SPC and MEMNET, is almost 0.05, a difference that seems to have some importance for an application. The CI at 95\% for this difference indicates that it could even go as far as 0.08 accuracy, but also that it could be as low as 0.013, a much more negligible value. These observations could lead to the conclusion that BERT~SPC is potentially a substantial improvement over MEMNET, but that further comparisons on independent data would be welcome, especially as MEMNET has fewer parameters. This analysis is not as favorable to the BERT models as the one conducted by \newcite{Song2019}, which was based only on the leaderboard, but it should be kept in mind that these authors analyzed several datasets and that the scores obtained are not perfectly identical.

On the other hand, the data collected clearly argue in favor of AEN~BERT and BERT~SPC compared to TD~LSTM since the CI lower limit is greater than 0.045.

\begin{table*}[!h]
\begin{center}
\begin{tabular}{llrrrrrrrrrrrr}
\toprule
\multicolumn{2}{c}{Condition} & \multicolumn{2}{c}{Correlation} & \multicolumn{3}{c}{p-value} & \multicolumn{2}{c}{Lower CL} & \multicolumn{2}{c}{Difference} & \multicolumn{2}{c}{Upper CL}\\
C1 & C2 & C1 & C2 & Min & Max & \#sig & Min & Max & Min & Max & Min & Max \\ \midrule
Full & ¬FC & .802 & .802 & .0315 & .9779 & 3 & -.010 & .001 & -.005	& .005 & -.001 & .011 \\ 
Full & ¬CNN & .802 & .791 & .0032 & .2930 & 7 & -.005 & .007 & .006	& .018	& .017 & .030 \\ 
Full & ¬LE & .802 & .710 & .0001 & .0001 & 20 & .062 & .081 & .084	& .107 & .111 & .140 \\ 
¬FC & ¬CNN & .802 & .791 & .0020 & .3341 & 5 & -.005 & .007 & .006	& .017 & .018 & .029 \\ 
¬FC & ¬LE & .802 & .710 & .0001 & .0001 & 20 & .062 & .083 & .084	& .108 & .110 & .140 \\ 
¬CNN & ¬LE & .791 & .710 & .0001 & .0001 & 20 & .042 & .066 & .069	& .097	& .101 & .133 \\ \bottomrule
\end{tabular}
\caption{Results for Joy of the twenty 10-fold CV for the ablation procedure applied to LE-PC-DNN with from left to right the names of the compared conditions, the respective correlations with the gold standard, the minimum and maximum p-value and the number of significant comparisons ($\alpha = .05$) for the Fisher-Pitman tests, and the minimum and maximum values of the lower limit of the CIs (CL), of the difference in correlation between the two conditions and of the upper limit of the CIs.}
\end{center}
\end{table*}

Figure 2 might give the impression that all the CIs are actually roughly the same length, reducing the interest of using them. If indeed four out of ten CIs have a length between 0.0689 and 0.0706, the largest length (0.0783 for BERT~SPC and TD~LSTM) is 125\% of the smallest (0.0627 for ATAE~LSTM and MEMNET). More generally, while there is a strong relationship between the conclusions of the tests (i.e, the a, b and c groupings) and those of the CIs, the crucial point is that the CIs make it possible to qualify the conclusions in relation to the simple significant vs. not significant dichotomy. For example, Figure 2 shows that there is a significant difference between MEMNET and TD~LSTM and not between ATAE~LSTM and TD~LSTM, but the lower CI limits are (almost) identical and therefore the significant difference should be put into perspective.

\subsection{Study 2: LE-PC-DNN for EmoInt}
The second case study illustrates the potential usefulness of CIs when performing K-fold CV. It is based on the WASSA'17 EmoInt shared task for which systems had to estimate the emotion intensity of tweets for four emotions (anger, fear, joy and sadness) on a real-valued scale \cite{Mohammad2017}. The performance was measured by means of the Pearson correlation coefficient. In this framework, \cite{Kulshreshtha2018} proposed \textit{LE-PC-DNN}, an improved version of the system ranked first in the challenge. I have reproduced their study using the code they provided at \url{https://github.com/Pranav-Goel/Neural\_Emotion\_Intensity\_Prediction} \cite{Goel}. In their paper, they report an ablation analysis of the performance of their system on the evaluation set by removing one by one a parallel connected component from their LE-PC-DNN architecture. 

I have reproduced this ablation analysis on the evaluation set using a K-fold CV procedure, with $K = 10$, and performed twenty different 10-fold CVs, obtained by varying the random seed. It is therefore assumed here that a researcher compares two systems on the basis of only one\footnote{There is thus no reason to apply a Bonferroni correction here, even more so because it seems difficult to justify it in the case of CVs \cite{BEST2020}.} of the many possible K-fold CVs, twenty in this case study.

To save space, only the results for Joy are shown in Table 1, but the full results are available in the downloadable materials. In three of the six comparisons, the permutation tests report statistically significant differences in all CVs and the CIs show that these differences seem indeed important, being at least 0.069. 

In the other three comparisons, three to seven of the CVs gave rise to statistically significant differences according to the Fisher-Pitman permutation test, but the others did not. Depending on the CV performed, the researcher's conclusion in terms of statistical significance would therefore be different. 

On the other hand, for all these comparisons, the observed difference is small ($\le 0.018$) and the maximum lower limits of the CIs are extremely small ($\le 0.007$). It follows that none of the CVs assessed on the basis of the CIs argues for a difference which seems sufficiently important. This analysis qualifies the conclusions of the original study, as it is far from clear that all components play an actual positive role in overall performance.

This case study shows that researchers who rely on the leaderboard or on a significance test applied to the evaluation set alone, as is very often the case, will most likely obtain unreliable conclusions. They can of course proceed as here by performing a large number of CVs, but it will be much simpler and less resource-intensive to calculate the CIs.
 
\section{Conclusion}
The two case studies presented above indicate that it is easy to obtain a CI for the difference between two systems and that it provides useful information when interpreting the results. However, this procedure is seldom used in NLP as evidenced for example by the absence of this term in the recent \textit{Synthesis Lectures on Human Language Technologies} by \cite{Dror2020} on \textit{Statistical Significance Testing for Natural Language Processing}, despite the close vicinity of these two approaches.

When developing a system or performing an ablation analysis, all the necessary data to compute the CIs are available. Computing them to compare systems proposed by different researchers is easy for the organizers of a challenge. These CIs could be calculated by default on platforms like Codalab, and on Multi-Task Benchmarks like GLUE, SuperGLUE or XTREME \cite{wang-etal-2018-glue,pmlr-v119-hu20b}. By simply adding the confidence limits to each score, everyone could get a clearer picture of the superiority of one system over others. 

However, calculating these CIs is much more difficult for individual researchers when they want to compare their system to the SOTA ones. This could become easier if it is recommended that, when a paper applies a system to a public dataset, the predictions for the evaluation set are made freely available to everyone, for instance in hubs of ready-to-use datasets for machine learning models \cite{lhoest-etal-2021-datasets}.

An important characteristic of the CI approach compared to the other ways of presenting learderboard results is that, for each task, it is up to the researchers or the community to decide what difference is sufficiently important. This forces researchers to go beyond the numbers and encourages them to take other parameters into account. For instance, a difference should be seen as more or less important depending on the resources necessary to obtain the performance \cite{Dodge2019}. Likewise, CIs including zero should argue for the most interesting system according to factors other than accuracy.

\section{Acknowledgements}

The author is a Research Associate of the Fonds de la Recherche Scientifique (FRS-FNRS). He would like to thank the reviewers for their very constructive comments.

\section{Bibliographical References}\label{reference}

\bibliographystyle{lrec2022-bib}
\bibliography{mylrec2022}

\end{document}